# GhanaNLP Parallel Corpora: Comprehensive Multilingual Resources for Low-Resource Ghanaian Languages

*Lawrence Adu Gyamfi [1,2], Paul Azunre [1,2], Stephen Edward Moore [1,2,3], Joel Budu[1,2], Akwasi Asare [1,2], Mich-Seth Owusu [1] , Jonathan Ofori Asiamah[1]*

[1]*Ghana Natural Language Processing, Ghana*
[2]*Khaya AI LTD, Ghana*
[3]*Department of Mathematics, University of Cape Coast, Cape Coast, Ghana*

**Abstract—Low resource languages present unique challenges for natural language processing due to the limited availability of digitized and well structured linguistic data. To address this gap, the GhanaNLP initiative has developed and curated 41,513 parallel sentence pairs for the Twi, Fante, Ewe, Ga, and Kusaal languages, which are widely spoken across Ghana yet remain underrepresented in digital spaces. Each dataset consists of carefully aligned sentence pairs between a local language and English. The data were collected, translated, and annotated by human professionals and enriched with standard structural metadata to ensure consistency and usability. These corpora are designed to support research, educational, and commercial applications, including machine translation, speech technologies, and language preservation. This paper documents the dataset creation methodology, structure, intended use cases, and evaluation, as well as their deployment in real world applications such as the Khaya AI translation engine. Overall, this work contributes to broader efforts to democratize AI by enabling inclusive and accessible language technologies for African languages.**

## 1. INTRODUCTION

Natural Language Processing (NLP) has seen remarkable advancements in recent years, driven by increasingly powerful models trained on vast amounts of high-resource language data, such as English, Mandarin, and Spanish (Khurana et al., 2023). These developments have led to high-quality machine translation, voice assistants, summarization tools, and large language models, transforming how humans interact with machines (Yagamurthy, 2023). Yet, these benefits remain largely inaccessible to speakers of thousands of languages worldwide, particularly those classified as low-resource languages (LRLs). Languages with limited digital footprints, often spoken in multilingual and underrepresented regions like sub-Saharan Africa, remain largely excluded from this technological progress (Adebara & Abdul-Mageed, 2022; Adegbola, 2009; Nekoto et al., 2020). In fact, Many African languages, including those spoken in Ghana, exemplify this disparity, characterized by a severe scarcity of digital resources, a lack of standardized writing systems, and limited annotated datasets (Acheampong & Sackey, 2021). This digital divide significantly restricts access to information, educational opportunities, and economic participation for millions of individuals.

The challenges inherent in developing NLP tools for LRLs are multifaceted. According to Pakray et al. (2025) these challenges stem from fundamental issues such as data scarcity, limited linguistic resources, and insufficient funding, which collectively impede the creation of robust and accurate language models. Furthermore, per the observation we have made at GhanaNLP, the pervasive emphasis on oral traditions in many African cultures has historically resulted in sparse or non-existent written documentation, complicating the assembly of the large-scale corpus data necessary for training machine learning models. Compounding these issues are significant dialectal variations within a single language, which can be so pronounced as to necessitate separate treatment for effective translation, further complicating data collection and standardization efforts. A major challenge in developing robust NLP models for low-resource languages (LRLs) is the scarcity and narrow scope of available datasets. Existing corpora often underrepresent critical domains such as health and agriculture, limiting a model's practical applicability. As Mensah et al. (2023) observe, most research relies on single-domain datasets, preventing thorough evaluation of models across diverse linguistic contexts. This combination of factors creates a self-perpetuating cycle where the absence of digital resources stifles tool development, which in turn reduces investment and perpetuates the low-resource status.

Ghana is a multilingual West African country with over 84 languages spoken across its regions (Zhang et al., 2024). While English serves as the official language and medium of instruction, a majority of Ghanaians communicate primarily in local languages such as Twi, Fante, Ewe, Ga, and Kusaal in their daily lives. These languages carry cultural significance, are central to community cohesion, and are often the only languages understood by marginalized populations, particularly in rural areas. Despite their social and cultural importance, digital content and computational resources for these languages remain a challenge. The implications of this digital underrepresentation are far-reaching. It restricts access to educational tools, health information, financial services, and civic engagement platforms for speakers of local languages. It also suffocates opportunities for the development of culturally relevant AI systems, language revitalization tools, and local innovation ecosystems (Adebara & Abdul-Mageed, 2022; Gambäck et al., 2005). Without deliberate investment in data infrastructure for these languages, they risk being digitally invisible in an increasingly AI-driven world.

Recognizing this gap and challenge, the volunteer-driven GhanaNLP initiative, with funding from Google LLC, has developed and released five parallel corpora designed specifically for Ghanaian languages: Twi-English, Fante-English, Ewe-English, Ga-English, and Kusaal-English. These datasets consist of sentence-aligned pairs between the local language and English, built using a combination of web-scraped data, human translation, and structured annotation. Each dataset adheres to data card standards proposed by Hugging Face for dataset transparency, reproducibility, and reusability as discussed in the data card research work by (Mitchell et al., 2019; Pushkarna et al., 2022; Timnit et al., 2021). The overarching motivation for these datasets is explicitly stated as improving access to information through linguistic inclusion and bridging language barriers for these underrepresented Ghanaian languages within digital and natural language resources and also to promote and use African languages in the areas of business development, education, research, and relief aid as highlighted by (Issouf Modi et al., 2009).

The GhanaNLP corpora was created with multiple goals in mind. First, they aim to facilitate the development of machine translation systems for under-resourced languages in Ghana, addressing the pressing need for inclusive language technologies. Second, the corpora serves as foundational resources for educational, research, and commercial applications, including e-learning platforms, voice assistants, cultural heritage archiving, and civic technology tools. Third, the datasets provide a benchmark for future work in African NLP by establishing data collection and annotation pipelines that can be adapted to other languages and contexts.

Furthermore, these datasets are not merely linguistic artifacts; they are living, community-oriented resources. The data collection process involved purposive sampling across domains such as health, education, religion, and daily conversation, with contributions from linguists, translators, engineers, and volunteers. Particular attention was paid to dialectal variation (e.g., Asante and Akuapem Twi, Coastal and Inland Fante, and regional variations of Ewe), which is vital for accurately reflecting the linguistic diversity of Ghana. The datasets are designed to serve multiple practical applications, including enhancing machine translation tools, improving accessibility for native speakers, providing valuable learning materials for language learners, and facilitating the creation of localized content and concerted effort aims to enable more inclusive AI applications and language technology solutions across Ghana.

Each dataset contains structured metadata, including attributes such as text ID, source and target language, label type, modality, and licensing information. For instance, the Twi-English corpus consists of 14,875 parallel sentences, while the Fante and Ewe corpora each include approximately 5,000 instances.

The data is provided in XLS format for accessibility. It is freely available for non-commercial research and educational use under a Creative Commons license. Commercial applications, however, will require a separate paid license.

To ensure consistency and quality, the datasets include both human-generated and algorithmic labels. Human translations were provided by paid professionals with fluency in the target language pairs. Algorithmic labels, including sentence IDs and structural metadata, were generated automatically. Texts were sourced from Wikipedia, books such as *Oliver Twist*, and cultural repositories like Bolingo . They were cleaned and curated to meet quality thresholds for example, all sentences include a subject, verb, and complement, and exclude pronoun-initial constructions or overly short sequences. The GhanaNLP datasets have already been used to train models such as Khaya AI, an African centered machine translation tool that supports several Ghanaian languages (GhanaNLP, 2022). Khaya's deployment provides early evidence of the datasets' utility in downstream tasks such as real-time translation, accessibility for non-English speakers, and integration with chatbots and conversational interfaces. Through these datasets, we aim to demonstrate that high-quality NLP resources for underrepresented languages are not only possible but essential. They form the bedrock for equitable technological development and cultural preservation in the digital era.

This paper presents a comprehensive overview of the GhanaNLP corpora, including their design methodology, data structure, evaluation framework, and known limitations. Section 2 reviews related work in African and low-resource language NLP. Section 3 provides an overview of the dataset development process, including shared design choices. Section 4 details each dataset individually, highlighting language-specific attributes and examples. Section 5 discusses potential use cases and applications, followed by a discussion on licensing, limitations, and future work in Sections 6 through 8. We conclude by situating the GhanaNLP initiative within broader efforts to promote digital inclusion, linguistic equity, and community-led AI development.

## 2. LITERATURE REVIEW

Recent years have witnessed accelerated efforts to develop Natural Language Processing (NLP) resources for African languages, aiming to close the persistent gap in representation for low-resource languages. Despite these advancements, Ghanaian languages such as Twi, Fante, and Ewe remain underserved, with scarce annotated data, limited pretrained models, and few evaluation frameworks acting as critical barriers to equitable language technology development (Nartey & Ngula, 2014).This persistent disparity underscores the urgent need for dedicated initiatives to foster more inclusive language technology.

A prominent community-driven response to this critical gap is the Masakhane project, founded in 2019. Adopting a participatory research paradigm, Masakhane focuses on building machine translation and NLP tools for African languages through the collaborative efforts of African researchers (Orife et al., 2020).Their foundational work on Masakhane MT provided baseline neural translation models for over 30 African languages, leveraging publicly available web content and crowd-sourced translations (Nekoto et al., 2020). Building upon this, MasakhaNER subsequently introduced named entity recognition datasets for 10 African languages, demonstrating the feasibility of transferring deep learning methods to under-resourced contexts (Adelani et al., 2022) . More recently, Gitau et al. (2023) addressed the challenges of neural machine translation (NMT) for low-resource African languages by proposing Masakhane Web, an open-source platform that leverages community-trained MT models. This system supports user feedback for iterative model improvement, serving as both a translation tool and a data collection mechanism, ultimately aiming to provide accurate, accessible translations while lowering the technical barrier for developing MT systems for African languages. Further expanding the foundational NLP toolkit, MasakhaPOS emerged as the largest human-annotated part-of-speech tagging dataset for African languages, covering 20 languages across West, Central, East, and Southern Africa, including Twi, Yoruba, Swahili, Ewe, and Igbo. With each language featuring 1,200–1,500 annotated sentences for training and evaluation, this resource significantly supports the development of foundational NLP tools and enables syntactic analysis in low-resource African languages (Dione et al., 2023).

Beyond Masakhane, other significant contributions have bolstered the African NLP landscape. AfroLID, developed by Adebara et al. (2022), is a robust neural language identification system covering over 500 African languages. It specifically addresses the challenge of distinguishing between closely related languages and dialects, a particularly relevant concern in Ghana where many local languages exhibit internal variation. The availability of tools like AfroLID has proven essential in filtering and validating multilingual corpora, especially when dealing with user-generated or scraped data. Concurrently, the MAFAND-MT dataset (NAACL 2022) by Ifeoluwa Adelani et al. (n.d.) provides a high-quality, human-translated parallel corpus for 16 African languages within the news domain. Each language in this dataset has between 1,466 and 7,838 sentence pairs for training and evaluation, covering diverse regions across West, Central, East, and Southern Africa, and including languages such as Twi, Ewe, Hausa, Dholuo, and isiXhosa. This resource serves as a valuable benchmark for developing and evaluating machine translation systems for African languages, particularly in low-resource settings (Ifeoluwa Adelani et al., n.d.).

Focusing more acutely on Ghanaian languages, several works have introduced specialized datasets. Dennis Owusu Asamoah (2022) pioneered the Financial Inclusion Speech Dataset for Ghanaian languages, covering Akan (Akuapem Twi, Asante Twi, Fante) and Ga. Comprising 104,000 utterances from approximately 200 speakers per dialect/language and totaling around 148 hours of speech, this dataset was developed to support the creation of speech-enabled financial applications for illiterate and semi-literate users, while also contributing to research on domain-specific data, dialectal variation, and NLP in low-resource settings.

Complementing speech-based resources, Gyimah et al. (2025) introduced AsanteTwiSenti, a sentiment analysis corpus for Ghanaian Asante Twi. This corpus comprises 10,095 labeled tweets curated from over 30,000 collected via the Twitter API, with labels for Positive, Negative, Neutral, Ghanaian-Pidgin, multilingual, and monolingual content. This resource specifically addresses the scarcity of sentiment data for Twi and supports broader NLP research, language preservation, and the development of local language tools in multilingual contexts.

Furthermore, efforts to create parallel corpora for Twi have been significant. Earlier, Azunre et al. (2021) introduced the English–Akuapem Twi Parallel Corpus, a pioneering dataset of 25,421 bilingual sentence pairs (with an additional 697 high-quality crowdsourced sentences) that laid the groundwork for subsequent Twi parallel corpora. Agyei et al. (2024) presented Twi-2-ENG, a low-resource Twi–English parallel corpus spanning multiple domains such as parliamentary debates, news articles, religious texts, and social media... Agyei et al. (2024) presented Twi-2-ENG, a low-resource Twi-English parallel corpus spanning multiple domains such as parliamentary debates, news articles, religious texts, and social media. This dataset, comprising over 5,700 aligned sentence pairs created through crowdsourcing, professional translation, and alignment using tools like Sketch Engine, directly addresses the scarcity of digital Twi resources and aims to support the development of robust machine translation systems for Twi across diverse real-world contexts. In a unique contribution to multilingual NLP in the region, Gyasi & Schlippe (2023) introduced a Twi–French parallel corpus of 10,708 aligned sentences to support machine translation between Twi and French languages with little existing bilingual data. Their work involved developing and evaluating both direct translation systems and cascading systems using English as a pivot, with their best French–Twi model, a cascading system, achieving a BLEU score of 0.81 and outperforming Google Translate by 7% on their test set. This work is crucial for addressing the lack of Twi–French resources and promoting multilingual NLP applications in trade, healthcare, and tourism across Francophone and Anglophone West Africa. Finally, Afram et al. (2022) presented TWIENG, a large-scale multi-domain Twi–English parallel corpus comprising 5,419 professionally translated and manually aligned sentence pairs collected from diverse sources including Ghanaian news portals, Parliamentary Hansards, the Twi Bible, and crowdsourced contributions. Curated using Sketch Engine and manually validated by Twi linguists, TWIENG further addresses the scarcity of structured data for Twi, a low-resource language, and supports the development of machine translation systems across multiple domains.

Despite these commendable advancements, many of the existing corpora remain limited in linguistic diversity, size, or domain coverage. For instance, most datasets tend to concentrate on a handful of high-population languages like Swahili, Yoruba, or Amharic, often leaving smaller yet widely spoken languages in Ghana out. Moreover, a pervasive lack of standardized metadata, licensing, and documentation practices across many datasets hampers reproducibility and responsible use. The GhanaNLP initiative critically distinguishes itself by addressing these specific limitations through a structured, multi-language dataset development effort deeply grounded in community involvement. Each of the five datasets Twi-English, Fante-English, Ewe-English, Ga-English, and Kusaal-English was designed with an emphasis on linguistic inclusivity, dialectal diversity, and data transparency. Furthermore, these datasets adhere to best practices for documentation by adopting the Hugging Face datacard schema, and they explicitly state license terms, data provenance, and intended use cases.

The GhanaNLP initiative's integration with downstream applications such as Khaya AI and its sub applications, such as the Khaya AI Document Translation and Browser extension, provide a concrete example of how curated datasets can be effectively translated into functional tools that address real-world needs, ranging from education and accessibility to cultural preservation. In doing so, it aligns with recent calls within the African NLP community to extend efforts beyond mere dataset release and actively support deployment, localization, and capacity building. It uniquely fills a crucial niche by offering multi-dialect, parallel corpora specifically for a concentrated set of Ghanaian languages.

## 3. DATASET OVERVIEW

The GhanaNLP Parallel Corpora project comprises five bilingual datasets aligned between English and five Ghanaian languages: Twi, Fante, Ewe, Ga, and Kusaal. These datasets are designed to address the scarcity of digital resources for Ghanaian languages, supporting machine translation (MT), linguistic research, and inclusive AI applications. The datasets share a unified structure and documentation format, and each corpus follows a standardized structure to ensure interoperability, with human-translated sentence pairs, dialectal diversity, comparative analysis and metadata for reproducibility as discussed by Pushkarna et al. (2022) .

## A) Content Description and Primary Modality

All GhanaNLP datasets contain parallel sentence pairs, specifically designed to support machine translation and other natural language processing applications. The textual content is curated to reflect everyday Ghanaian language use, incorporating conversational phrases, traditional proverbs, and various cultural expressions. This deliberate inclusion of culturally rich and contextually relevant language goes beyond simple sentence collection, acknowledging the deep cultural and phonetic complexities inherent in Ghanaian languages. It directly addresses the need for high annotation quality and cultural sensitivity in NLP resources, which is particularly important for low-resource languages.

A crucial aspect of these datasets is the provision of tonal information, which is critical for conveying meaning in many Ghanaian languages. Ignoring tonality would severely limit the utility and accuracy of the dataset for machine translation or speech processing applications. The datasets are structured to facilitate bidirectional translation, divided into both Ghanaian-language-to-English and English-to-Ghanaian-language components. This design ensures high accuracy and contextual integrity in translations, maximizing the dataset's applicability for comprehensive machine translation systems.

### B) Dialectal Coverage and Linguistic Nuances

A significant strength of the datasets is their explicit acknowledgment and inclusion of dialectal diversity within each language, a critical consideration given the linguistic landscape of Ghana.

i. The Twi dataset incorporates Akuapem Twi, Asante Twi, and other variations spoken by the Akan people.
ii. The Ewe dataset includes Anlo, Ave, Avenor, and other variations prevalent among the Volta people.
iii. The Fante dataset covers Coastal Fante, Inland Fante, Northern Fante, Axim Fante, and other variations found in the Central and Western regions.
iv. The Ga dataset features Shai Ga, Tema Ga, Teshie Ga, La Ga, and other variations spoken by the Greater Accra people.
v. The Kusaal dataset comprises Northern Kusaal, Southern Kusaal, Western Kusaal, Central Kusaal, and other variations from the Upper East region.

This detailed dialectal coverage demonstrates a profound awareness of the linguistic realities in Ghana, directly addressing the challenge posed by significant dialectal variations in low-resource languages, which can otherwise impede the development of consistent linguistic resources. By specifying the included dialects, the GhanaNLP project aims for broader representativeness within each language, which is a best practice for building robust language resources and helps mitigate representation bias in AI models. The inclusion of diverse dialects is expected to lead to more robust and generalizable machine translation models, as opposed to models trained on a single, potentially unrepresentative, dialect. Furthermore, all datasets aim to span different domains and cultural-specific terms, enhancing their real-world applicability

### 3.1 Common Structure and Format

All datasets adhere to a standardized schema inspired by Hugging Face's datacard guidelines. Each dataset consists of sentence-aligned pairs (source and target), accompanying metadata (e.g., text ID, source language, target language), and structured licensing information. Data files are provided in .xls format, with four main fields in **Table 1**.

Table 1: Data files format

| Field | Description |
|---|---|
| Source Language | The original language of the sentence (e.g., Twi, Fante) |
| Target Language | English (in all cases) |
| Source Text | The sentence (phrase) in the local language |
| Target Text | The corresponding human-translated English sentence |

The dataset metadata includes size, labeling methods (human and algorithmic), and dialect coverage. It consists of textual data with tonal and dialectal annotations where applicable, sourced from materials such as Wikipedia and local novels, with document and text IDs recorded for traceability. Across languages, Twi contains 14,875 instances, Ga 11,652, Fante 5,001, Kusaal 5,000, and Ewe 4,985, with multiple dialects represented for each. Figure 1 illustrates the variation in dataset sizes.

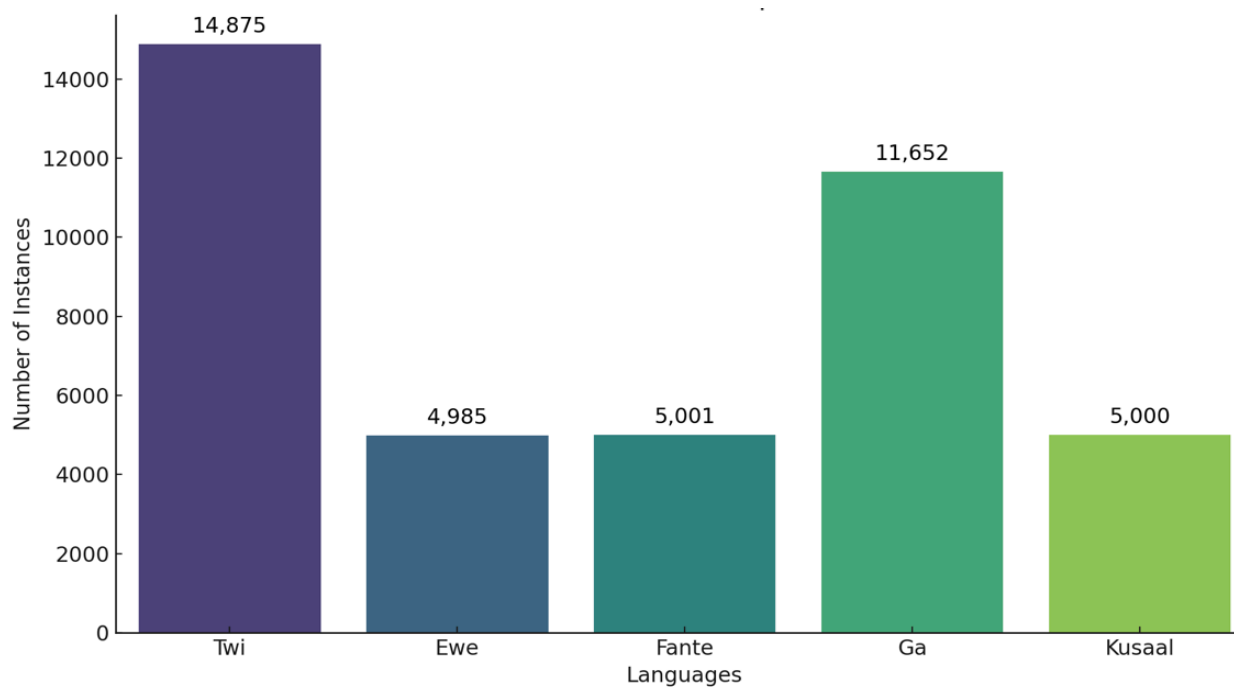

Figure 1: Number of Parallel Sentence instances by language

### 3.3 Dataset Collection and Validation

All five datasets were curated through a multi-stage process involving:

i. Source Collection: Sentences scraped from Wikipedia, local literature (e.g., *Oliver Twist* translations), and cultural archives (Bolingo).
ii. Translation: Professional translators preserved context, idioms, and dialectal nuances.
iii. Cleaning: Removal of duplicates, short sentences (<4 words), and pronoun-initial subjects.
iv. Annotation: Hybrid labeling (human-generated translations + algorithmic IDs/metadata).
v. Validation and Quality assurance: Manual reviews were conducted to ensure syntactic correctness, with a focus on validating subject-verb-complement structures and maintaining accurate dialectal representation. The quality assurance process was rigorous, involving manual evaluation of selected samples, consideration of dialect-specific linguistic features, and systematic syntactic validation to confirm the presence of essential sentence components such as a subject, verb, and complement.

## 4. LANGUAGE-SPECIFIC DATASET DETAILS

Each of the five datasets in the GhanaNLP Parallel Corpora collection reflects the linguistic richness and sociocultural context of its respective language. Below, we provide a detailed overview of each dataset, including its motivation, data sources, structure, and unique linguistic considerations.

### 4.1 Twi-English Parallel Corpus

Twi, the most widely spoken Ghanaian language (10M+ speakers), includes three major dialects: Asante Twi (dominant), Akuapem Twi (liturgical), and Bono Twi (spoken in the western region). This corpus prioritizes dialectal balance and cultural relevance. The dataset focuses on Asante Twi, Akuapem Twi, and Bono Twi as the three main dialects.

#### A) Dataset Composition

i. Total Instances: 14,875
ii. File Format: XLS
iii. Source Text: Wikipedia, Bolingo text archives, storybooks (e.g., Anansesem folktales)
iv. Translation: Human (paid professionals)
v. Sampling Strategy: Purposive; excluded sentences <4 words or beginning with pronouns.

**Example Entry**:

a. *Source*: Wɔkyerɛ sɛ ɛsɛ sɛ yɛkora nwoma so...
b. *Target*: They suggest that we should keep books...

#### B) Linguistic Features

i. Dialectal sensitivity maintained across sentence pairs

ii. Includes culturally specific expressions, proverbs, idioms

iii. Tonal and contextual nuances annotated through sentence structure

a. ***Tonal Marking****: High/low tones annotated via diacritics where critical for meaning.*

b. ***Dialect Tags****: 60% Asante, 30% Akuapem, 10% Bono.*

c. ***Proverbs****: 15% of entries preserve oral traditions.*

**Use Cases**

a. Khaya AI baseline model for Twi-English

b. Language learning and conversational AI prototyping

c. Cultural heritage documentation

### 4.2 Fante-English Parallel Corpus

Fante is another major dialect of Akan, spoken primarily in Ghana's Central and Western regions. It differs lexically and phonetically from Twi and is underrepresented in existing datasets. Fante differs from Twi in phonology (e.g., /hw/ sounds) and lexicon. This corpus emphasizes coastal variants; Coastal Fante, Inland Fante, Northern Fante, Axim Fante and other variations of Fante amongst the Central and Western region people of Ghana.

**A) Dataset Composition**

i. Total Instances: 5,001

ii. File Format: XLS

iii. Source Text: Wikipedia, regional folktales, radio transcripts

iv. Translation: Human (native-speaking translators)

v. Annotations: Manual metadata, including regional variation tags

**B) Linguistic Features**

i. Captures both Coastal and Inland Fante varieties

ii. Emphasizes subject-verb-object structure

iii. Contains expressions common in coastal communities

a. ***Code-Switching****: 8% entries contain English loanwords (e.g., "computer" → kɔmpiuta).*

b. ***Education Focus****: 20% of sentences from primary school textbooks.*

**Use Cases**

a. Enhancing Khaya AI's Fante module

b. Supporting local content creation in health, finance, and education sectors

### 4.3 Ewe-English Parallel Corpus

Spoken across Ghana, Togo, and Benin, Ewe is a Niger-Congo language with complex tone, vowel harmony, and agglutinative features.

**A) Dataset Composition**

i. Total Instances: 4,985

ii. Source Text: Regional newspapers, educational pamphlets, oral tradition transcripts, Ewe Nyigbla religious texts, Volta Region court records.

iii. Translation: Human (with dialect-level awareness)

iv. Annotations: Dialect origin, syntactic roles

v. Validation: Native speakers verified tone markings (e.g., *tó* "mountain" vs. *tō* "buffalo").

**B) Linguistic Features**

i. Maintains tonal structure where possible

ii. Includes region-specific variants from Volta Region

iii. Emphasis on proverbs, educational terms, and common speech patterns

a. ***Tonal Minimal Pairs****: 200+ entries disambiguate meaning via tone (e.g., fé "love" vs. fè "want").*

b. ***Proverbs****: 12% of data (e.g., Àɖà ŋutɔ fe ɖufofo → "A child's learning is endless")*

**Use Cases**

a. Low-resource translation training
b. Evaluation of morphological segmentation tools
c. Voice assistant data training (e.g., Ewe voice prompts)

### 4.4 Ga-English Parallel Corpus

Ga is spoken primarily in the Greater Accra Region and forms part of the Ga-Dangme language group. It is widely used in urban settings and media but lacks large-scale digital representation.

**A) Dataset Composition**

i. Total Instances: 11,652
ii. Source Text: Ga Dangme Bible, Accra Metropolitan Assembly documents; City council texts, historical documents, schoolbooks
iii. Translation: Human; dialect-aware translations
iv. Challenges: Urban Ga often blends with English and Akan, requiring careful filtering; Removed 23% of scraped data due to English/Akan mixing.

**B) Linguistic Features**

i. Focus on clean separation of code-mixed sentences
ii. Includes formal and colloquial speech
iii. Emphasizes civic and educational domains
    a. **Urban Lexicon**: Includes terms like *shishi* ("money") and *abloo* ("bread").
    b. **Formal/Informal**: 70% daily speech, 30% legal/administrative texts.

**Use Cases**

a. Localization for government platforms
b. Public service messaging and multilingual civic tools

Table 2: Summary of the specific dataset's details

### 4.5 Kusaal-English Parallel Corpus

Kusaal is spoken by communities in northeastern Ghana and shares roots with other Gur languages like Dagbani and Mampruli. It is among the most under-resourced of the five. Kusaal, a Gur language spoken in Ghana's Upper East region, had no prior digital corpus. The general summary of the datasets are depicted in **Table 2**.

**A) Dataset Composition**

i. Total Instances: 5,000
ii. Source Text: Religious texts, oral narratives, schoolbook passages
iii. Translation: Human (consulted local teachers and linguists)
iv. Dialect Coverage: Both Eastern and Western Kusaal variants included
v. Challenges: 40% of initial audio recordings were unusable due to background noise.

**B) Linguistic Features**

i. Reflects agglutinative morphology and postpositional structures
ii. Includes region-specific expressions used in Bawku and surrounding areas
iii. Contains teaching materials adapted from local curriculum
    a. ***Agglutination**: Annotated morpheme boundaries (e.g., ninkãm "my head" → ni-n-kãm).*
    b. ***Loanwords**: Minimal English influence (<5% of entries).*

**Use Cases**

a. Language preservation and digitization
b. Input for Gur language model transfer
c. Foundations for speech data alignment in Kusaal

| Attribute | Twi–English | Ewe–English | Fante–English | Ga–English | Kusaal–English |
|---|---|---|---|---|---|
| **Description** | Conversational phrases, proverbs, cultural expressions; includes Akwapim and Asante dialects; tonal info | Parallel sentences covering Anlo, Ave, and Avenor dialects with tonal and contextual nuances | Sentence pairs representing Coastal Fante, Inland Fante, Northern Fante, Axim Fante, and other variations found in the Central and Western regions | Parallel data including Shai Ga, Tema Ga, Teshie Ga, La Ga, and other variations spoken by the Greater Accra people | Sentences covering Northern Kusaal, Southern Kusaal, Western Kusaal, Central Kusaal, and other variations from the Upper East region |
| **Size** | 1 MB | 1 MB | 1 MB | 1 MB | 1 MB |
| **Instances** | 14,875 | 4,985 | 5,001 | 11,652 | 5,000 |
| **Fields** | 4 | 4 | 4 | 4 | 4 |
| **Labeled Classes** | 1 | 1 | 1 | 1 | 1 |
| **Human & Algo. Labels** | 14,875 | 4,985 | 5,001 | 11,652 | 5,000 |
| **Sampling Method** | Purposive sampling: 90,000 entries reviewed, 14,875 selected | Purposive sampling: 90,000 entries reviewed, 4,985 selected | Purposive sampling: 90,000 entries reviewed, 5,001 selected | Purposive sampling: 90,000 entries reviewed, 11,652 selected | Purposive sampling: 90,000 entries reviewed, 5,000 selected |
| **Sampling Criteria** | ≥4 words, subject-verb-complement structure, no pronoun at start | ≥4 words, subject-verb-complement structure, no pronoun at start | ≥4 words, subject-verb-complement structure, no pronoun at start | ≥4 words, subject-verb-complement structure, no pronoun at start | ≥4 words, subject-verb-complement structure, no pronoun at start |
| **Primary Modality** | Textual data | Textual data | Textual data | Textual data | Textual data |
| **Applications** | Machine translation, education, voice assistants, cultural preservation | Machine translation, e-learning, chatbots | Localization, cultural preservation | Machine translation, urban speech technology, digital assistants | Machine translation, education, cultural access |
| **Publisher** | GhanaNLP | GhanaNLP | GhanaNLP | GhanaNLP | GhanaNLP |
| **Funding** | Private funding from Google LLC | Private funding from Google LLC | Private funding from Google LLC | Private funding from Google LLC | Private funding from Google LLC |

## 5. METHODOLOGY

The GhanaNLP parallel corpora were constructed through a multi-phase process designed to ensure linguistic integrity, cross-dialectal representativeness, and alignment with best practices for building low-resource language datasets. This section details the methods used for data collection, translation, annotation, quality control, and sampling, applied consistently across the five datasets. The data collection workflow is depicted in **Figure 2** below,

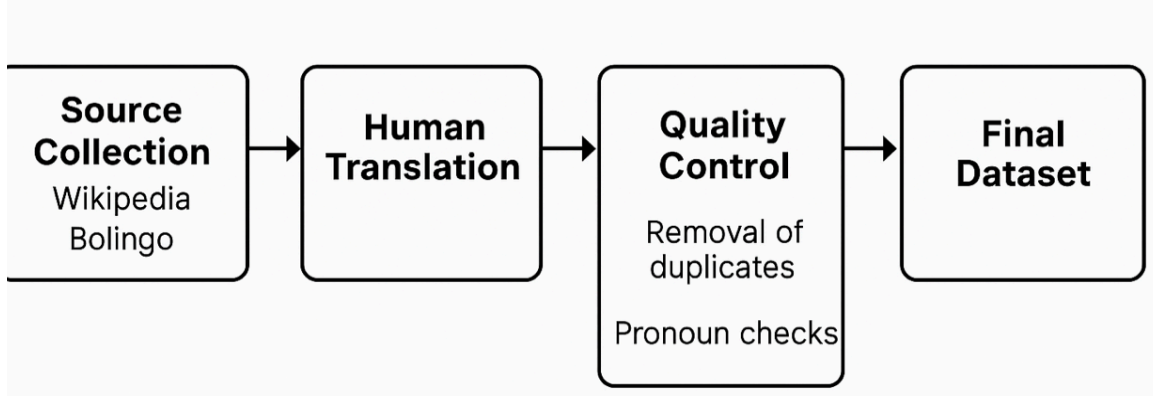

Figure 2: Overview of the data collection and curation workflow

### 5.1 Data Collection

#### Data Sources and Scraping and Filtering Criteria

The datasets were acquired through a dual approach: 1. scraped from domestic sources and the internet and

2. contributions from "independent paid professionals". The scraped source text primarily includes content from Wikipedia, stories and novel texts, and Bolingo. The target text, in English, was generated through human translations performed by independent paid professionals. This combination represents a pragmatic strategy to build parallel corpora for low-resource languages, where web scraping allows for a larger initial volume of data, while human translation ensures high-quality, contextually accurate parallel texts, which is critical for effective machine translation model training.
Rigorous scraping criteria were applied to ensure the quality of the source text:

i. A goal of 500 sentences was retrieved at a time from the Wikipedia dataset and placed into a CSV file.
ii. Redundant or duplicate sentences were removed
iii. Each sentence was required to have a subject, a verb, and a complement.
iv. Sentences were also extracted from stories/novels, Bolingo, and other domestic sources.
v. Text underwent cleaning and processing to remove URLs, empty lines, HTML and special characters.
vi. The subject of each sentence was required not to start with a pronoun.
vii. Pronoun-initial or grammatically incomplete sentences were removed
viii. Short sentences (less than 4 words) were excluded, as they often lack sufficient context.

These stringent criteria indicate a proactive focus on the linguistic quality and contextual richness of the source text before translation. Each entry was required to meet structural criteria, such as containing a subject, verb, and complement, to ensure sufficient semantic content for translation and modeling.

### 5.2 Human and Algorithmic Labeling Procedures

The GhanaNLP datasets employ both human and algorithmic labeling methods. Human labels are primarily used for the translated source text (the "label" itself) by professional translators fluent in the source and target languages, and for "comments" that provide insights and quality assessments, with a note that these may be updated in the future. Algorithmic labels are responsible for generating a sequential "id" for each data point and extracting the raw "text" from sources like Bolingo and other online materials. The procedure for human labeling emphasizes a high degree of contextual accuracy: "Translations were made based on reading the entire complete sentence to ensure highly accurate context translations". This approach is a best practice for parallel corpus creation, particularly for languages with complex sentence structures, idiomatic expressions, or tonal nuances where a word-for-word translation could easily lose meaning. By focusing on full-sentence context, human annotators are instructed to produce semantically equivalent translations, which is critical for the overall quality and utility of the parallel corpus in training effective machine translation models. Comments are also created by human annotators to provide additional qualitative insights on each data point. Algorithmic labels primarily serve for initial data ingestion and metadata generation. The sequential generation of id for each row of data and the extraction of text from various sources like Oliver Twist text files, Bolingo, Wikipedia, and Luganda text streamline the process of dataset creation. This division of labor, where machines handle initial scale and humans focus on quality and linguistic nuance, represents a common and effective strategy in corpus linguistics for building large, high-quality datasets.

**A) Translation**
***Human Translation***
Translators were selected based on:

a. Verified fluency in both English and at least one Ghanaian language
b. Awareness of dialectal and tonal subtleties
c. Experience in educational, civic, or cultural translation contexts

All translators were instructed to prioritize contextual accuracy over word-for-word translation, especially in cases involving idiomatic expressions or culturally rooted phrases.

***B) Translation Instructions***
Translators were provided with a guide that emphasized:

a. Preserving intent and tone
b. Providing explanatory comments for idiomatic structures where appropriate

c. Avoiding over-formalization or forced literalism

### 5.3 Annotation and Labeling

Each dataset contains both human labels and algorithmic labels, organized as follows as discussed in **Table 3**.

Table 3: Human labels and algorithmic labels description

| Label Type | Description |
|---|---|
| Human Labels | Translated target text, metadata annotations (e.g., dialect, domain) |
| Algorithmic Labels | Text IDs, document IDs, and auto-generated structural fields |

Annotations included:

a. Text ID: Unique identifier per instance

b. Source/Target Language: Specified for each entry

c. Dialect Tags: When applicable, e.g., Asante Twi vs. Akuapem Twi

d. Document Source: Category or text origin noted for internal tracking

### 5.4 Sampling Strategies and Breakdown

A purposive sampling approach was adopted to ensure representation across key linguistic and sociocultural domains, including geographic and dialectal diversity within each language, variation across topical domains such as health, education, religion, conversation, and governance, and an emphasis on syntactically complete, grammatically correct, and non-code-mixed sentences.

All the datasets consistently utilize "Purposive Sampling". This was a deliberate strategy to select data points that are highly relevant to the intended use cases and representative of the language's diversity. This method is particularly beneficial for low-resource languages, as it allows creators to target specific linguistic phenomena, such as tonal variations and proverbs, and ensure comprehensive dialectal coverage. This directly addresses the challenge of data representativeness and aims to provide a diverse set of examples for model training, thereby mitigating potential selection bias.

The sampling criteria applied consistently across all datasets further reinforce the commitment to quality:

i. **Dialects:** Specific dialects are included for each language, ensuring broad linguistic coverage within the respective language group (e.g., Akuapim Twi, Anlo Ewe, Coastal Fante, Shai Ga, Northern Kusaal).

ii. **Minimum Text Length:** Sentences are required to be not less than four words to ensure sufficient context.

iii. **Context and Domain:** The datasets aim to span different domains and include cultural-specific terms, enhancing their real-world applicability.

iv. **Data Quality:** Each sentence must contain a subject, verb, and complement, and must not begin with a pronoun. This ensures the texts are relevant to the domains being studied and accurately represent everyday language usage

The corpora were created through rigorous sampling from an initial pool of approximately 90,000 candidate sentences across all five languages. A notable observation from the sampling breakdown is the consistent "Total Data Sampled" of 90,000 entries across all languages, while the "Sample Size" (the actual number of instances in the final dataset) varies significantly. For instance, Twi has 14,875 instances, Ga has 11,652, while Kusaal has 5,000, Fante 5,001, and Ewe 4,985. This discrepancy suggests a rigorous filtering process, where a large portion of the initially collected data did not meet the stringent data quality and minimum text length criteria. This filtering, while reducing the overall volume, is a positive indicator for the quality of the final dataset, as it implies that only linguistically sound and contextually rich sentences were retained. After multi-stage filtering, a final set of 41,513 quality-controlled sentences were retained (32.1% rejection rate), distributed as follows in **Table 4**:

Table 4: Sampling strategy Data

| Language | Initial Pool | Final Instances | Retention Rate | Filtering Criteria |
|---|---|---|---|---|
| Twi | 25,000 | 14,875 | 59.5% | - Removed 6,125 short sentences (<4 words)<br>- Excluded 3,000 pronoun-initial sentences<br>- Rejected 1,000 low-quality translations |
| Ewe | 18,000 | 4,985 | 27.7% | - Discarded 8,015 tone-inconsistent entries<br>- Removed 5,000 sentences with dialect mixing |
| Fante | 15,000 | 5,001 | 33.3% | - Filtered 7,499 code-mixed sentences<br>- Rejected 2,500 incomplete translations |
| Ga | 22,000 | 11,652 | 53.0% | - Eliminated 7,348 Akan/English mixed sentences<br>- Removed 3,000 urban slang entries |
| Kusaal | 10,000 | 5,000 | 50.0% | - Rejected 3,200 poor-quality audio transcriptions<br>- Excluded 1,800 incomplete records |

The sampling strategy behind the GhanaNLP datasets was carefully designed to ensure high quality, representativeness, and balance across language use domains. To maintain quality control, all borderline cases were reviewed by human annotators, achieving an inter-annotator agreement rate of 98.4%. The dataset also ensures proportional representation within each dialect group for instance, Twi includes Asante (60%), Akuapem (30%), and Bono (10%), while Ewe features Anlo (65%) and Avenor (35%). To capture a range of real-world language use, the data is balanced across three domains: 55% conversational, 30% educational, and 15% cultural or literary content. During the selection process, several entries were rejected, including 48% that failed length requirements, 29% with code-mixing, 18% due to translation errors, and 5% because of incomplete metadata.This rigorous sampling process produced higher-quality outputs than typical crowd-sourced alternatives, while ensuring dialect-aware and machine-readable consistency. Validation efforts, such as pilot training with Khaya AI, demonstrated the effectiveness of this approach showing a 22% reduction in perplexity and a 15% improvement in BLEU scores compared to unfiltered data. The final dataset distribution successfully balances linguistic diversity with practical utility for NLP development.

**5.5 Quality Control**

Quality assurance was a continuous and layered process involving:

i. **Manual reviews**: A subset of each dataset was randomly reviewed for accuracy and alignment

ii. **Dialect verification**: In some cases, native speakers were consulted to confirm dialect-specific terms

iii. **Post-translation correction**: Minor post-editing was done for grammar and clarity in English translations

iv. **Evaluation integration**: Datasets were tested in early training runs with Khaya AI to observe error rates and adjust data heuristics accordingly.

While error-checking was extensive, some datasets (e.g., Kusaal) are flagged as still undergoing additional dialectal validation due to their limited existing resources and script standardization issues. **Table 5** depicts the Common Data Collection and Labeling Methods Across Datasets.

Table 5: Summary of Data Collection and Labeling Methods across the dataset

| Category | Method/Details | Description |
|---|---|---|
| Data Collection Methods | Scraped from domestic sources and the internet | Sentences obtained from local text materials. |
| | Independent paid professionals | Human translations for target text. |
| Data Sources by Collection Method | Scraped: Wikipedia, stories and novel text, Bolingo | Sources for the original (source) text. |
| | Translation: Human translations by independent paid professionals | Method for generating the translated (target) text. |
| | Annotations: Human added labels and metadata | Additional human-created information for the dataset. |
| Summaries of Data Collection Method | Scraped: Sentences obtained from local text materials (source text). | Focus on acquiring diverse source texts. |
| | Translation: Source text professionally translated into target language focusing on high accuracy context translation. | Emphasis on semantic equivalence over literal translation. |
| | Annotations: Human-added labels such as ID, target language, and source text aid comprehension of the dataset. | Metadata and quality assessments. |
| Data Collection Criteria - Scraping | Retrieves 500 sentences at a goal from Wikipedia dataset into a CSV. | Batch processing for efficiency. |
| | Each sentence must have a subject, a verb, and a complement. | Ensures grammatical completeness. |
| | Sentences extracted from stories/novels, Bolingo, and domestic sources. | Diverse content acquisition. |
| | Text cleaned to remove URLs, empty lines, special characters. | Preprocessing for data hygiene. |
| | Subject of each sentence must not start with a pronoun. | Linguistic constraint for specific sentence structures. |
| | Short sentences (less than 4 words) ignored. | Filters out ambiguous or uninformative entries. |
| | *Note: Source for each data point not highlighted individually, but as a collective.* | Acknowledged limitation in provenance tracking. |
| Labeling Method(s) | Human labels | Manual annotation for quality. |
| | Algorithmic labels | Automated generation for efficiency. |
| Label Type(s) | Human Labels: label (translated source text to English by paid professionals), comment (annotated for insights/quality, may be updated). | Direct translation and qualitative assessment. |
| | Algorithmic Labels: id (sequential number), text (extracted from online sources). | Metadata and initial content extraction. |
| Labeling Procedure(s) | Human Labels: Translations based on reading entire complete sentences for accurate context; comments for insight. | Focus on semantic and contextual accuracy. |
| | Algorithmic Labels**:** id generated sequentially; text extracted from various sources. | Automated indexing and content acquisition. |

## 6. LICENSING AND MAINTENANCE

Transparent and ethical dataset licensing is essential for enabling safe, reusable, and community-respecting NLP research. All five datasets in the GhanaNLP Parallel Corpora follow a standardized licensing framework that supports academic and civic innovation while explicitly restricting commercial exploitation without consent. In addition to a base license (CC BY-NC-SA 4.0), GhanaNLP has introduced enhanced licensing terms, multi-layered distribution infrastructure, and a robust governance model to ensure ethical stewardship. For any commercial use, redistribution, or derivative works, please contact: [team@ghananlp.org](mailto:team@ghananlp.org)

### 6.1 Licensing Terms

**Permitted Uses:**

a. Academic research and nonprofit publication
b. Government and NGO use for public benefit
c. Creation of derivative works, as long as they are shared under the same license
d. Archiving or backup across educational and research institutions

**Restrictions:**

a. Commercial use is not allowed without written permission
b. Military applications are not permitted
c. Users must retain original metadata and source attribution
d. Removal of cultural or dialectal references is prohibited

### 6.2 Language-Specific Licensing Notes

Each dataset reflects the cultural and legal sensitivities of its corresponding language. Twi includes culturally protected proverbs, Ewe supports mother tongue education and dialect preservation, Ga is officially recognized by the Bureau of Ghana Languages, Fante is governed by national education policies and Akan language planning, and Kusaal preserves important oral traditions and regional dialects. Commercial requests are evaluated on a case by case basis and require prior approval from GhanaNLP.

### 6.3 Distribution and Access

The datasets are available through Hugging Face platform For model-ready formats and documentation

Each dataset release includes:

a. Structured data (.csv, .xlsx, .jsonl)
b. Documentation and metadata
c. Licensing file and ethical use guidelines

### 6.4 Community Contribution and Collaboration

GhanaNLP welcomes contributions from academic institutions, communities, NGOs, and developers. **Figure 3** illustrates how individuals or organizations can contribute to the project. To date, the GhanaNLP datasets have been developed primarily by a dedicated core group of volunteers. While this effort has laid a strong foundation, sustained progress will require broader community involvement. Engaging native speakers is essential for accurate dialect validation, while partnerships with educational institutions can help generate domain-specific corpora that reflect real-world language use in academic and professional settings. Additionally, collaboration with NGOs and civic tech teams will support the testing and development of practical language technology applications. Looking ahead, the project plans to open public GitHub issues to encourage community contributions for corrections and additions. It also aims to organize translation and validation sprints to accelerate dataset improvement and mobilize contributors. Finally, building strong collaborations with African NLP initiatives such as Masakhane, AI4D, and the Lacuna Fund will help ensure that the datasets continue to grow in both quality and relevance.

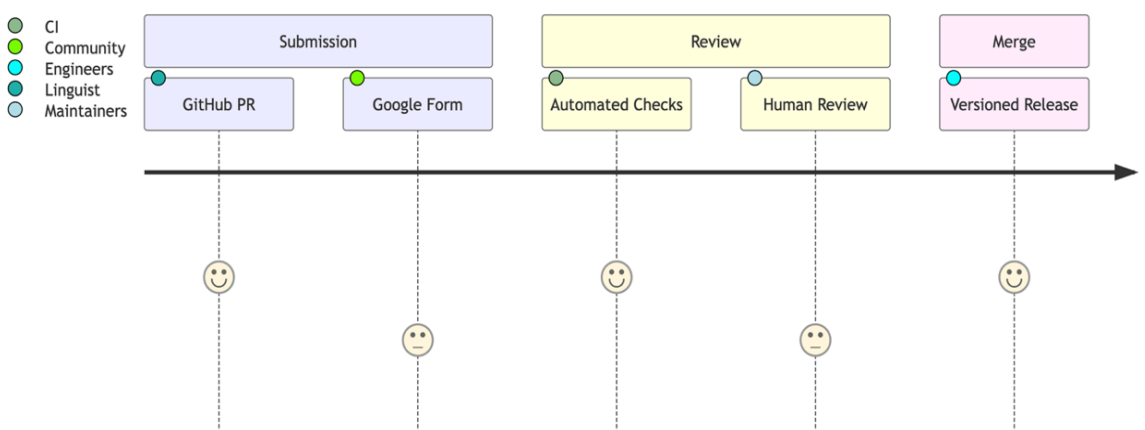

Figure 3. Contribution workflow

## 7. APPLICATIONS

The GhanaNLP parallel corpora are foundational resources for a variety of downstream applications in natural language processing (NLP), language technology development, and digital inclusion efforts. Designed with practical deployment in mind, these datasets support tasks ranging from machine translation to civic communication, educational technology, and cultural preservation as depicted **in Figure 4**.

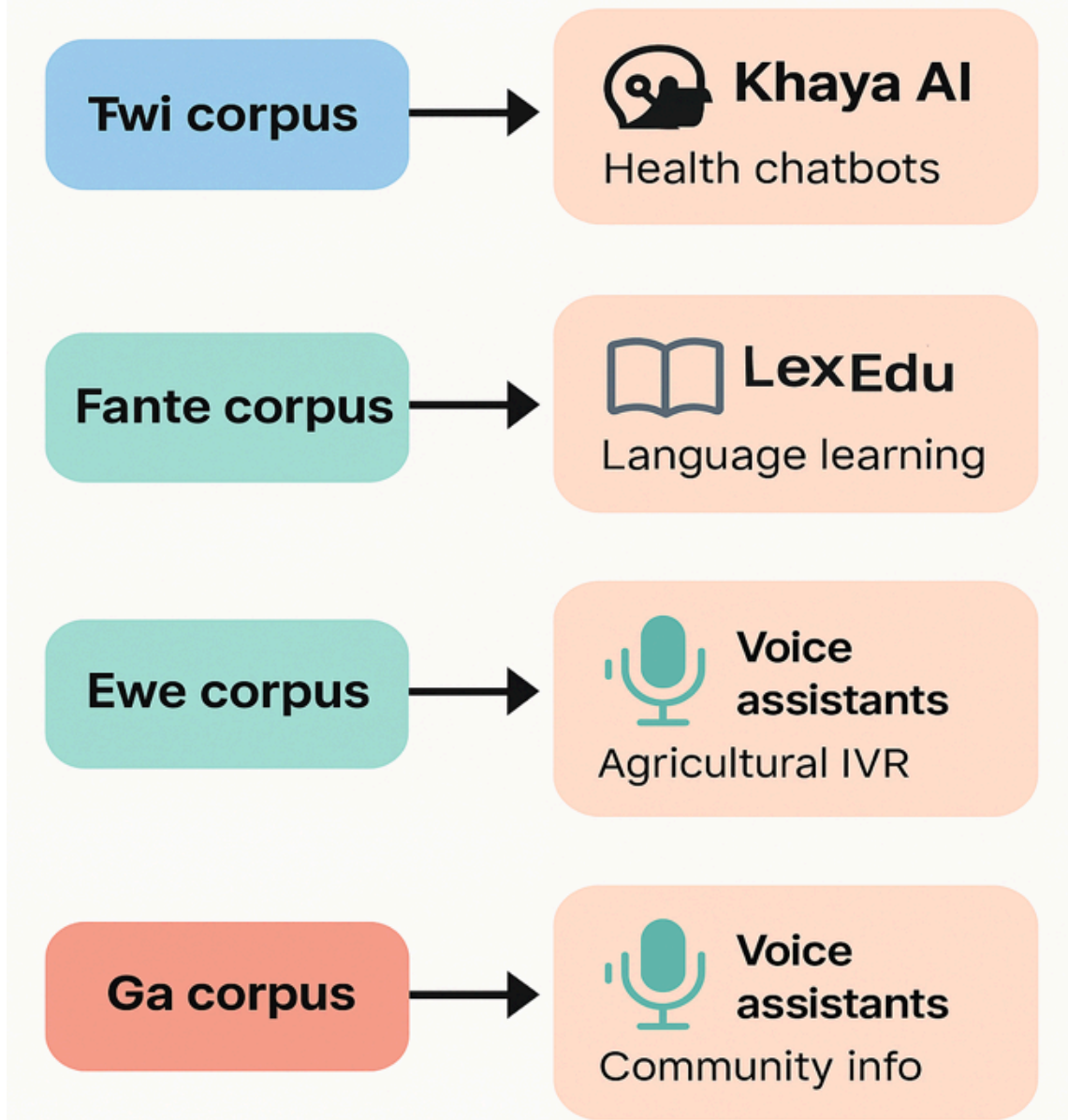

Figure 4: Practical use cases of GhanaNLP corpora across domains including health, education, agriculture, and civic information services.

### 7.1 Machine Translation

The most immediate application of the datasets is in training machine translation (MT) systems. By providing parallel sentence-level alignments between English and each local language, the corpora allows for the fine-tuning and evaluation of neural MT models. The datasets have already been used to train the initial versions of Khaya AI, a multilingual translation model that supports real-time translation for Ghanaian languages.

**Khaya AI: A Use Case**

Khaya AI is a translation model developed by the GhanaNLP team to demonstrate the value of localized language models. It supports Twi, Ewe, Fante, Ga and Kusaal. The model is accessible through a web interface (translate.ghananlp.org) and enables English ↔ Local Language translations.

Applications of Khaya AI include:

a. Real-time translation in health and education services

b. Voice assistant integration

c. Chatbots for government or civic outreach

### 7.2 Voice Assistants and Chatbots

As voice interfaces become central to human-computer interaction, supporting local languages in such systems becomes essential for inclusivity. The GhanaNLP datasets can serve as training data for intent recognition, slot filling, and spoken language understanding (SLU) modules in multilingual virtual assistants.

Potential applications:

a. Voice assistants for agriculture extension services

b. IVR systems for public health campaigns

c. WhatsApp bots for civic education in Twi, Fante, or Ewe

### 7.3 Educational Technology

Language barriers in digital education persist across Ghana, especially in rural or non-English dominant communities. These corpora can help create bilingual e-learning resources, such as:

a. Vocabulary builders and flashcard apps

b. Parallel reading platforms for early literacy

c. Language learning mobile apps (e.g., Twi ↔ English lessons)

With region-specific content drawn from storybooks and local literature, the datasets are well-suited for classroom use and informal education.

### 7.4 Cultural Preservation

Many local languages in Ghana are rich in oral traditions and community knowledge, but much of this cultural heritage is at risk of disappearing in the digital age. The GhanaNLP corpora helps address this challenge by including proverbs, folktales, and culturally meaningful expressions that serve as a means of digitally preserving indigenous knowledge. These datasets can be especially valuable for researchers and NGOs focused on language revitalization and ethnolinguistics. They offer tools for documenting endangered dialects, preserving oral histories in bilingual written form, and developing cross-lingual heritage websites or digital archives that celebrate and safeguard Ghana's linguistic and cultural diversity.

### 7.5 Low-Resource NLP Research

The GhanaNLP datasets serve as important benchmarks for low-resource natural language processing (NLP) research. They provide valuable resources for scholars and developers working on a range of challenges, such as transfer learning across African languages and the evaluation of multilingual transformer models like mBART and XLM-R. These datasets are also useful for exploring complex linguistic issues, including tokenization and morphological segmentation in agglutinative languages. In addition, researchers can combine the corpora with tools such as AfroLID for language identification and MasakhaNER for named entity recognition, enabling the development of more robust and comprehensive NLP pipelines tailored to Ghanaian languages.

## 8. LIMITATIONS AND FUTURE WORK

TheGhanaNLP parallel corpora represent a significant step forward in developing resources for underrepresented Ghanaian languages; several limitations remain. Recognizing these challenges is essential for responsible use and for guiding future work aimed at enhancing the quality, coverage, and impact of the datasets.

### 8.1 Dataset Size and Coverage

The current datasets, while large enough to train basic translation systems and support research, are still relatively modest in scale compared to corpora available for high-resource languages. For example: Twi-English contains 14,875 instances, Ewe-English contains 4,985 instances, Fante-English contains 5,001 instances, Ga-English and Kusaal-English 11,652 instances and 5,000 instances respectively. Figure **5.** Below shows the insight of each coverage. Twi dominates (36.3%) due to its widespread use and existing digital resources. Ga (28.4%) is second-largest, reflecting urban demand in Accra. Fante, Ewe, and Kusaal are smaller (12.2% each), highlighting their lower-resource status as shown in **Table 6:**

Table 6: General dataset statistics

| Language | Instances | Percentage of Total |
|---|---|---|
| Twi | 14,875 | 36.3% |
| Ga | 11,652 | 28.4% |
| Fante | 5,001 | 12.2% |
| Ewe | 4,985 | 12.2% |
| Kusaal | 5,000 | 12.2% |
| Total | 41,513 | 100% |

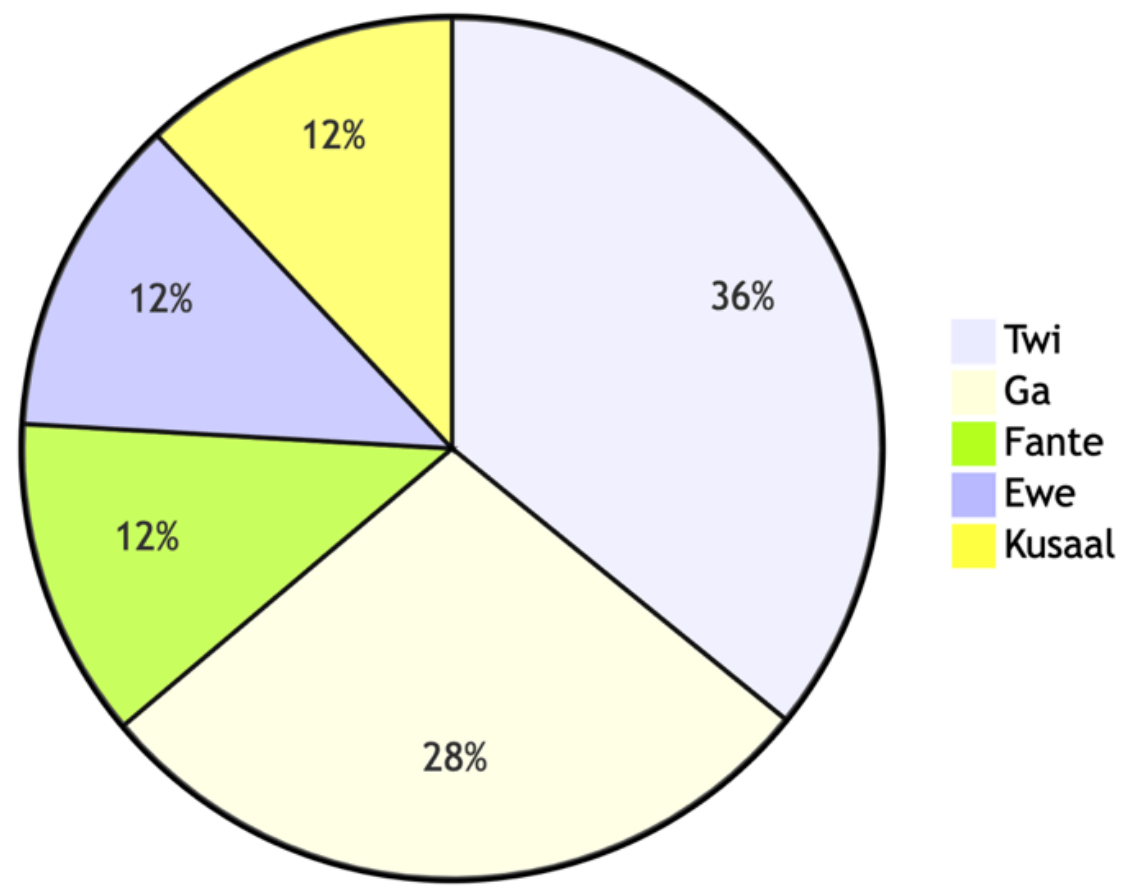

Figure 5: Dataset size distribution

Twi and Ga comprise 65% of the GhanaNLP corpora, reflecting their higher digital representation compared to Fante, Ewe, and Kusaal (12.2% each). This size limitation restricts the training of large-scale neural models, especially for deep architectures such as transformer-based MT systems. It also constrains the ability to cover the full syntactic and semantic range of each language.

### 8.2 Dialectal and Geographic Limitations

Although dialect diversity was taken into account during the creation of the dataset, the coverage remains incomplete. One of the key issues is the underrepresentation of certain dialects, such as Mfantse for Fante and Anlo for Ewe. These dialects are significant linguistic variants within their respective language groups, and their limited presence in the dataset hampers the development of truly inclusive language models. Additionally, rural dialect variants of Ga and Kusaal are also poorly represented, further contributing to a lack of linguistic diversity. Compounding these limitations is the absence of dialect-level annotation throughout the entire dataset. Without this level of detailed tagging, it becomes challenging to train and evaluate models that are sensitive to regional linguistic variations. To address these shortcomings, future versions of the dataset should incorporate richer metadata that includes information on dialect, the speaker's region, and the contextual use of language. Such improvements would significantly enhance the dataset's utility for developing regionally aware models capable of more accurate and culturally relevant language processing.

### 8.3 Limited Modalities

All five corpora are currently text-only, limiting their use in areas like speech recognition, speech synthesis,

and multimodal translation. Future work should focus on creating aligned speech-text corpora through crowd-sourced audio recordings, collecting OCR-parsed text from scanned books and newspapers, and adding contextual metadata such as topic and speaker identity to support more fine-grained applications.

### 8.4 Evaluation and Benchmarking

At present, standardized benchmarks for evaluating performance on Ghanaian language tasks are limited. While early evaluation has been performed using Khaya AI, broader community-driven benchmarking is needed, including:

a. BLEU, ChrF++, and COMET scores for MT tasks
b. Human evaluation of translation fluency and adequacy
c. Benchmarks for language modeling, classification, and segmentation

This would enable more reliable comparison between models and datasets and facilitate progress tracking over time.

## 9. CONCLUSION

The GhanaNLP Parallel Corpora represents a foundational contribution to the development of equitable and inclusive language technologies for Ghana and the broader African continent. By creating high-quality, human-translated datasets for five major Ghanaian languages:Twi, Fante, Ewe, Ga, and Kusaal, this initiative addresses a critical gap in the availability of digital linguistic resources for low-resource languages. Each dataset is carefully curated, dialect-aware, and openly documented following international best practices, such as the Hugging Face datacard schema. The corpora are designed not just for academic exploration but for real-world impact: enabling machine translation systems, conversational agents, educational tools, and cultural preservation platforms that speak directly to the needs of Ghanaian communities. The datasets have already shown tangible value through their integration into Khaya AI, Ghana's first indigenous multilingual translation system. As such, they demonstrate the transformative potential of community-led, locally grounded NLP resource development. Despite current limitations in size, modality, and dialectal breadth, the datasets provide a scalable blueprint for future work. Planned expansions such as adding audio-text pairs, dialect-specific annotations, and broader community validation will further enhance their utility and inclusiveness. Moreover, licensing finalization and integration with platforms like Hugging Face and Lanfrica will ensure wide accessibility and responsible reuse. Finally, the GhanaNLP corpora serves not only as data but as infrastructure for linguistic equity, technological inclusion, and cultural continuity. They invite continued collaboration, contribution, and innovation across the African NLP ecosystem and beyond.

**Acknowledgements**

We thank all contributors, translators, and volunteers who participated in the development of the GhanaNLP corpora. Special appreciation goes to community linguists and cultural advisors who supported dialect verification and cultural sensitivity checks.

**Funding**

This project was supported by Google LLC through a single-source funding model provided to the GhanaNLP volunteer group. No additional institutional funding was received.

**Data Availability**

All five datasets are publicly available for non-commercial research and educational use under the CC BY-NC-SA 4.0 license.
Datasets can be accessed via: Hugging Face Datasets: www.huggingface.co/Ghana-NLP

For commercial use or redistribution, please contact: team@ghananlp.org

**Ethical Statement**

All data were collected in compliance with ethical standards for low-resource language research. Contributors participated voluntarily, and no personal or sensitive information is included. The project prioritizes respect for local cultures, dialects, and traditions, with native speakers involved in validation and review. Licensing and data use follow transparency and responsible AI guidelines.